\documentclass[letterpaper, 10 pt, conference]{ieeeconf}  

\IEEEoverridecommandlockouts                              

\usepackage{cite}

\usepackage{amsmath,amssymb,amsfonts,graphicx,url,subcaption,float,booktabs,hyperref}
\usepackage[ruled,vlined]{algorithm2e}
\usepackage{graphicx, caption, booktabs, tabularx, multirow, array}
\usepackage{textcomp}
\usepackage[top=20.1mm, left=16.9mm, right=16.9mm, bottom=15.2mm]{geometry}
\usepackage[usenames,dvipsnames,table]{xcolor}
\graphicspath{{Sections/figs/}}

\usepackage{titlesec}
\titlespacing\section{0pt}{3pt plus 1pt minus 1pt}{3pt plus 1pt minus 1pt}
\titlespacing\subsection{0pt}{2pt plus 1pt minus 1pt}{2pt plus 1pt minus 1pt}
\titlespacing\subsubsection{0pt}{2pt plus 1pt minus 1pt}{1pt plus 1pt minus 1pt}
\usepackage[export]{adjustbox}

\makeatletter
\renewcommand{\thanks}[1]{\protected@xdef\@thanks{\@thanks
        \protect\footnotetext[0]{#1}}}
\makeatother

\begin{document}
\title{\LARGE \bf
Single-Eye View: Monocular Real-time Perception Package for Autonomous Driving
}

\author{Haixi Zhang$^{*1}$, Aiyinsi Zuo$^{*1}$, Zirui Li$^{*1}$, Chunshu Wu$^1$, Tong Geng$^2$, Zhiyao Duan$^2$}

\thanks{$^*$Equal Contribution}
\thanks{$^{1}$Haixi Zhang, Aiyinsi Zuo, Zirui Li, and Chunshu Wu are with the Department of Electrical and Computer Engineering, University of Rochester, Rochester, NY 14627, USA \\ {\tt\small \{azuo, zli133, hzh104\}@u.rochester.edu, cwu88@ur.rochester.edu}}%
        
\thanks{$^2$Zhiyao Duan and Tong Geng are with the Faculty of the Department of Electrical and Computer Engineering, University of Rochester, Rochester, NY 14627, USA \\{\tt\small\{zhiyao.duan,tong.geng\}@rochester.edu}%
}
\date{}
\maketitle
\thispagestyle{empty}
\begin{abstract}

Amidst the ascendency of camera based autonomous driving technology, sometimes effectiveness is overally focused with limited attention to computational demand. To address these issues, this paper introduces LRHPerception, a real-time monocular perception package for autonomous driving that uses single view monocular camera videos to produce interpretation of the surroundings. It innovatively fuses the computational efficiency of end-to-end learning with the comprehensive details intrinsic to local mapping methodologies: With significant enhancements across object tracking and prediction, road segmentation, and depth estimation encapsulated in a cohesive system, LRHPerception proficiently processes monocular image data, yielding a five-channel tensor comprising the original RGB, road segmentation, and pixel-level depth estimation channels, embellished with object detection and trajectory prediction. Empirical evaluations substantiate its superior performance, showcasing a single-GPU real-time processing rate of 29 FPS, a 555\% acceleration over the fastest mapping technique, attributed equally to modular enhancements and our unique amalgamation technique. The code is available at \href{https://github.com/EnisZuo/LRHPerception}{\textcolor{RoyalBlue}{LRHPerception}}.

\textbf{Index Terms---}
Computer Vision for Transportation, Deep Learning for Visual Perception, Visual Recognition
\end{abstract}

\setlength{\belowcaptionskip}{-16pt}\
\captionsetup[figure]{font={small}}
{\fontsize{9pt}{10pt}\selectfont
\begin{figure*}
  \centering
  \includegraphics[width=1\linewidth]{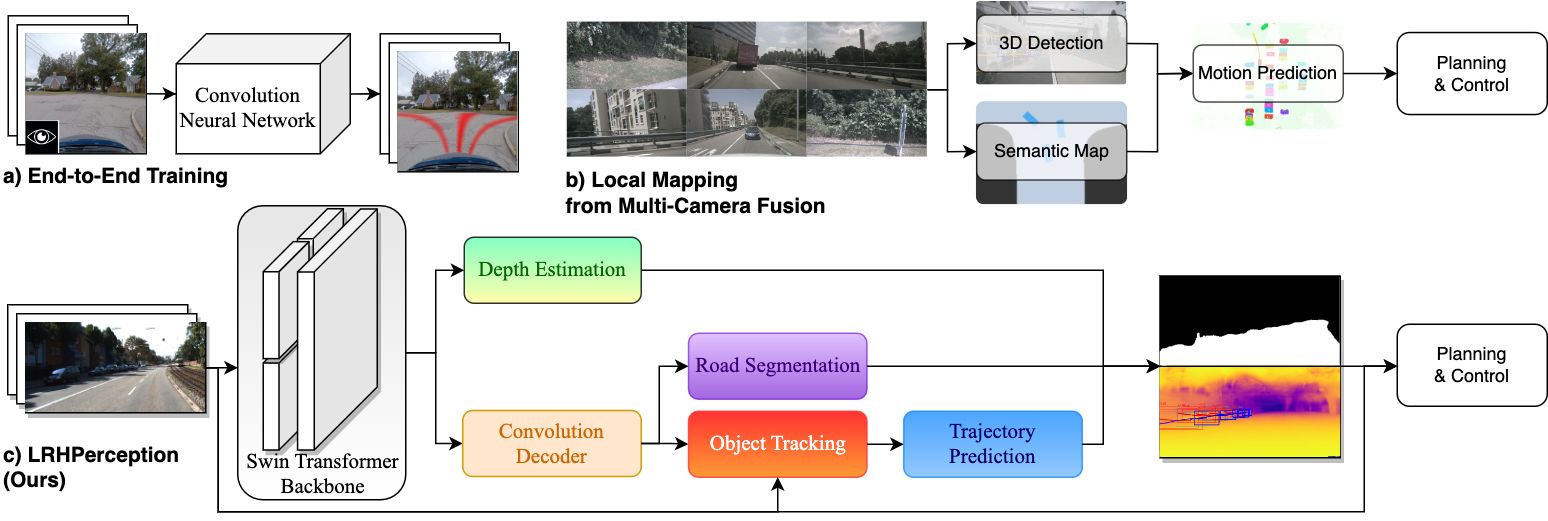}
  \caption{\textbf{Innovation and architecture blueprint} a) Paradigm of end-to-end solution b) Paradigm of camera-fusion for local map solution c) Paradigm of our LRHPerception package, extracts essences from monocular camera for cost-info trade-off.}
  \label{fig：first}
\end{figure*}}
\setlength{\belowcaptionskip}{0pt}

\vspace{-1em}
\section{Introduction}
In recent years, the domain of autonomous driving has witnessed significant progress, particularly in cost-effective camera-based technologies. Such research is also driven by the aspiration to mirror human driving, ultimately leading to safer and more understandable vehicles~\cite{wu2022human}. Predominantly, two methodologies have emerged: 1) end-to-end neural networks that interpret raw images to directly produce steering commands~\cite{bojarski2016end,hou2017fast}, and 2) multi-camera fusion techniques that generate a bird's eye view or 3D occupancy map for path planning~\cite{heng2019project,wang2022detr3d,chen2022learning}. While groundbreaking, they have their respective constraints.

End-to-end learning methods transform raw imagery into driving decisions, lauded for their computational prowess~\cite{hou2017fast}. Yet, they often face scrutiny due to limited interpretability and unpredictability from simplified input processing design~\cite{norden2019efficient}. These systems, despite performing admirably under training-similar conditions, might falter in unfamiliar, dynamic traffic scenarios~\cite{tampuu2020survey}, potentially leading to  unsafe decisions. Conversely, systems utilizing imagery from multiple cameras provide an encompassing view of the environment, aiding technicians in comprehension and system enhancement. Their computational demands, however, can sometimes impede real-time processing on standard hardware with single GPU, limiting practicality in live scenarios. 

Our paper presents a novel monocular real-time perception system for autonomous driving, addressing key challenges in this domain. Traditional approaches have extensively researched individual areas like road detection, pixel-depth estimation, object detection, and trajectory predictions. However, these domains have often been studied in isolation. We bridge this gap by not only enhancing each module but also introducing novel integration techniques that seamlessly fuse these components. By processing single-camera video feeds, our system, LRHPerception (\textbf{L}ow-cost, \textbf{R}eal-time, \textbf{H}igh Information richness), offers a robust blend of resource efficiency with an information-rich perception of the driving scenario.


LRHPerception processes RGB images, providing road segmentation, pixel-depth estimation, object detection, and trajectory predictions. Unique modules for each task integrate computationally efficient blocks and structures, ensuring comparable or superior accuracy. Integration involves shared backbones and skip feature map connections, reducing repetitive input data processing. This pioneering effort represents the first instance of amalgamating these modules into a comprehensive package. Our key contributions include:

1)	We introduce ``LRHPerception", a unified and pragmatic approach to autonomous perception that efficiently implements object tracking, trajectory prediction, road segmentation, and depth estimation, all derived from a monocular camera's video input. To our knowledge, this is the first work to integrate these modules into a cohesive package for real-time processing.

2)	We implement substantial innovation across each module of monocular image perception, consistently surpassing contemporary state-of-the-art benchmarks. These advancements not only contribute to faster processing speeds but also achieve either comparable or superior perception accuracy.

3) We present an integration technique that consolidates all modules within the package, facilitating information sharing to reduce redundant processing. This integration, combined with module-specific innovations, achieves a 555\% acceleration compared to the fastest local-mapping method.

4) The LRHPerception package constitutes the first block in the robotic pipeline of ``Perception-Cognition-Action" under our vision of creating a practical and efficient monocular-camera-based autonomous driving system.

\section{Related Works}
\subsection{End-to-End Training}
Contrary to traditional methods, which segregate processes like localization and mapping, planning, and control~\cite{pendleton2017perception}, end-to-end algorithms seek to unify these processes into a single learned model, directly translating raw sensory data into output control commands. This idea was first exemplified by the ALVINN system~\cite{pomerleau1988alvinn}, which employed a multilayer perceptron to learn the vehicle's steering direction. With the rise of convolutional neural networks (CNNs), the ability to learn deterministic~\cite{bojarski2016end} or probabilistic~\cite{amini2018spatial} driving commands from raw imagery has significantly evoloved, enabling sustained driving ~\cite{tampuu2020survey} and intricate lane-change maneuvers~\cite{7995938}.

Existing models often struggle to address the inherent ambiguity, commonly described as the ``black-box nature,''~\cite{norden2019efficient} in yielding steering possibilities, which can create obscure limitations and pose a potential safety risk. Moreover, they frequently neglect the importance of considering interactions with other traffic participants~\cite{pomerleau1988alvinn,amini2018spatial}, which can hinder performance, especially in complex and dynamic traffic scenarios. 

\subsection{Local Mapping via Multi-Camera Fusion}
The fusion of multi-camera data to enable simultaneous localization and mapping (SLAM) and thus construct and update maps of uncharted environments, has attracted substantial research attention~\cite{ heng2019project,chen2022learning,hu2023planning}. Its capacity to furnish a panoramic view of the surroundings enriches trajectory prediction and distance computation: Numerous methodologies have been proffered to accomplish multi-camera fusion to link with downstream tasks of object tracking and prediction~\cite{zhang2022beverse, gu2023vip3d}. Despite their remarkable results, the inherent complexity of the tasks precludes deployment on a singular, unified system~\cite{hu2023planning}. Specifically, current state-of-the-art models only reach a processing rate of less than 10 frames per second~\cite{hu2023planning}.

\setlength{\belowcaptionskip}{-14pt}
{\fontsize{9pt}{10pt}\selectfont
\begin{figure*}
  \centering
  \includegraphics[width=1\linewidth]{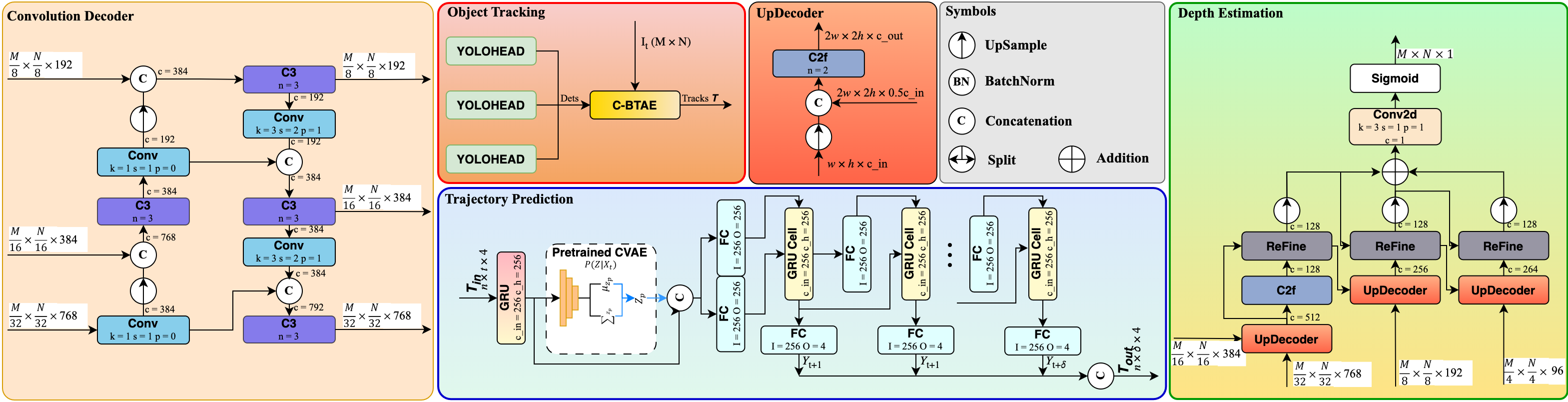}
  \vspace{-1.5em}
  \caption{\textbf{Granular Model Structure.1} Design of convolution decoder, object tracking, trajectory prediction, and depth estimation; magnify for details. BTAE mechanism in Algorithm 1. Remaining components  are shown in Fig. 3.}
  \vspace{-0.5em}
  \label{fig：second}
\end{figure*}}
\setlength{\belowcaptionskip}{0pt}

\section{Method: LRHPerception} 
We present a state-of-the-art autonomous perception package designed for monocular image inputs, striking an optimal balance among interpretability, information richness, and computational efficiency, ensuring real-time processing capabilities. Our package's architecture (see Fig.\ref{fig：first}) encompasses key functionalities of object tracking, trajectory prediction, road segmentation, and depth estimation. Unlike serial connections, we've integrated these modules within our model to facilitate information sharing, reducing redundant input processing.

For common feature recognition, we have chosen transformer backbones, particularly Swin Transformer~\cite{liu2021swin}. Acknowledged for its versatility, Swin Transformer outperforms convolutional backbones, notably in depth estimation~\cite{ming2021deep}. This backbone ingests an RGB image to generate feature-extracted maps $\Phi_{2^k}$ with stride size $2^k$, where $k \in \{2,3,4,5\}$.

The feature maps, namely $\{\Phi_4,\Phi_8,\Phi_{16},\Phi_{32}\}$, are routed to the aforementioned modules for subsequent tasks. Specifically, $\{\Phi_8,\Phi_{16},\Phi_{32}\}$ are relayed to the convolution decoder for simultaneous pixel segmentation and trajectory prediction. Meanwhile, all feature maps contribute to the Depth Former to synthesize a singular depth map layer. The final output of the LRHPerception package is a composite of the original RGB input $I$, segmentation $O$, depth map $D$, and a trajectory prediction overlay $\{Y_{t+1},...,Y_{t+\delta}\}$.

Such an integration yields significant computational savings. When detection, tracking, segmentation, and depth estimation are processed individually from the input, computation for three backbones is needed. In contrast, our architectural design leverages a shared backbone and two distinct feature extractions-- one from the convolution decoder and the other from the depth estimation module. This strategy lets us complete four tasks with only computation cost for one backbone. 

Beyond the computational efficiencies from the integration technique, intrinsic innovations within each functional domain also hold significant importance. These innovations secure efficiency improvements without compromising the efficacy of individual tasks, as detailed in the subsequent sections.

\subsection{Object Tracking}

Object tracking stands fundamental to autonomous driving, rooted deeply in object detection paradigms of computer vision~\cite{leon2021review}. A suite of detection algorithms, namely Faster R-CNN~\cite{girshick2015fast},  YOLO series~\cite{yolov8_ultralytics}, and more, are harnessed to bolster tracking performance.

Once detection on a single frame is formulated, the focus of most tracking techniques shifts towards establishing data association across frames. The Kalman filter (KF)\cite{kalman1960new} is a popular choice to anticipate tracklet locations and attain tracklets matching via location similarity~\cite{leon2021review}. Additional cues, such as camera movement and low-confidence boxes, have been factored in by various methods to achieve cutting-edge results~\cite{zhang2022bytetrack, hou2019vehicle, aharon2022bot}. Our module innovates on these cues to produce refined results for tracking robustness and safety:

Our \textbf{C-BYTE} (Camera-Calibrated BYTE) approach, depicted in Fig.\ref{fig：second}, is our contribution aimed at formulating tracking trajectories with diminished errors. Distinguished from the original BYTE method~\cite{zhang2022bytetrack} that hinges on bounding box correlations, our further strategy incorporates a camera movement correction mechanism between two adjacent frames to refine associations with immediate camera motion. This is particularly pertinent in scenarios of autonomous driving applications where vehicles are constantly moving.

For the processes in focus, consider an input video sequence. After processing to produce a set of bounding boxes from prior modules and Yolohead, the C-BYTE mechanism is activated. First, we perform Lucas-Kanade optical flow between current-time image $I_{t}$ and previous-time image $I_{t-1}$ to obtain $\mathcal{P}_t$, points with estimated positions in $I_t$ corresponding to the original keypoints in $I_{t-1}$\cite{sharmin2012optimal}. The generation of keypoints, $\mathcal{P}_{t-1}$, will be discussed later in this section. Lucas-Kanade method approximates optical flow within a local window around each keypoint, employing spatial and temporal image gradients to solve the flow equation. This least-squares fit computes motion vectors for each point, estimating keypoints' displacement from the previous to the current frame. Then, we calculate the affine matrix $\textbf{A}\in\mathbb{R}^{2\times3}$ using Random Sample Consensus (RANSAC) with previous and current time keypoints, $\mathcal{P}_{t-1}$ and $\mathcal{P}_{t}$\cite{derpanis2010overview}. In RANSAC algorithm, a minimal subset of point correspondences is randomly selected to estimate the affine transformation matrix. This matrix is iteratively refined through repeated trials, selecting the model with maximum inliers. The final matrix represents the optimal geometric transformation between frames, robust to outliers. 

\SetKwComment{Comment}{// }{ }

\begin{algorithm}
\DontPrintSemicolon
\SetKwInput{KwInput}{Input}
\SetKwInput{KwOutput}{Output}
\KwInput{Detected objects' bounding boxes at current time $\mathcal{O}$, existing tracks $\mathcal{T}$, current frame $I_t$, previous frame $I_{t-1}$, previous keypoints $\mathcal{P}_{t-1}$}
\KwOutput{Updated tracks $\mathcal{T}$}

\textcolor{ForestGreen}{/* Affine Matrix Application */\;
$\mathcal{P}_t \gets \text{LKOpticalFlow}(I_{t-1},I_{t},\mathcal{P}_{t-1})$\;
$\textbf{A} \gets \text{RANSAC}(\mathcal{P}_{t-1},\mathcal{P}_{t})$\;
$\mathcal{O}_{pred} \gets \text{Transform}(\text{KalmanFilter}(\mathcal{T}), A)$\;}
/* Primary Association */\;
$\text{Split } \mathcal{O} \text{ to } \mathcal{O}_{\text{high}},\mathcal{O}_{\text{low}} \text{ based on detection score}$\;
$\textbf{C} \gets mIOU(\mathcal{O}_{pred},\mathcal{O}_{high})$\;
\text{Associate $\mathcal{T}$ and $\mathcal{O}_{high}$ using $\textbf{C}$ }\;
/* Secondary Association */\;
$\mathcal{O}_{\text{high-remain}} \gets$ remaining object boxes from $\mathcal{O}_{\text{high}}$ \;
$\mathcal{O}_{\text{pred-remain}} \gets$ remaining objects boxes from $\mathcal{O}_{pred}$ \;
$\mathcal{T}_{\text{remain}} \gets$ remaining tracks from $\mathcal{T}$ \;
$\mathcal{T}_{\text{re-remain}} \gets$ remaining tracks from $\mathcal{T}_{\text{remain}}$ \;
$\textbf{C} \gets mIOU(\mathcal{O}_{pred-remain},\mathcal{O}_{low})$\;
$\text{Associate } \mathcal{T}_{remain} \text{ and } \mathcal{O}_{low} \text{ using } \textbf{C}$ \;
/* Remove unassociated and initialize new tracks */\;
$\mathcal{T} \gets \mathcal{T} \setminus \mathcal{T}_{\text{re-remain}} $ \;
\For{each $o$ in $\mathcal{O}_{\text{high-remain}}$}{
    $\mathcal{T} \gets \mathcal{T} \cup \{o\}$ \;
}
\textcolor{ForestGreen}{/* Update previous frame keypoints */\;
$\mathcal{P}_{t-1} \gets \emptyset$\;
$I_l \gets \text{LaplacianOperation}(I_t)$\;
\For{each \text{point} $(x,y)$ in $I_l$ }{
\uIf{$I_l[x,y] > \theta_{th}$}{
$\mathcal{P}_{t-1} \cup \{(x,y)\}$\;}
}}
\Return $\mathcal{T}$ \;
\caption{Pseudo-code of Key Steps in C-BYTE}
\end{algorithm}

Now, we separate and apply transformations with the affine matrix following \cite{Siciliano2008-gg,aharon2022bot}: $\textbf{A}\in\mathbb{R}^{2\times3} = \left[ R_{2 \times 2} | O_{2 \times 1} \right]$, where $R$ and $O$ represent Cartesian coordinate rotation and displacement matrices respectively. Then we apply these matrices separately to the anticipation of the bounding boxes' positions and velocities ($x,y,w,h\text{; }v_x,v_y,v_w,v_h$) of all current tracks $\mathcal{T}$ in current-time image $I_t$ derived from the Kalman filter: For $(x,y)$, apply displacement transformation:
$\begin{bmatrix}
x' \\
y'
\end{bmatrix} = O + \begin{bmatrix}
x \\
y\end{bmatrix}$.
For every two variables in these eight parameters, apply rotational transformation similarly to:
$\begin{bmatrix}
x' \\
y'
\end{bmatrix} = R \begin{bmatrix}
x \\
y\end{bmatrix}$.

Then, we split the original detection boxes into two lists based on their detection scores: $\mathcal{O}_{\text{high}}$ and $\mathcal{O}_{\text{low}}$. Next, we perform a two-step association following \cite{zhang2022bytetrack}:
    Primary Association- $\mathcal{O}_{\text{high}}$ are matched with $\mathcal{T}$ using transformed predictions $\mathcal{O}_{pred}$ from the Kalman filter.
    Secondary Association- We link $\mathcal{O}_{\text{low}}$ with the remaining tracks, denoted as $\mathcal{T}_{\text{remain}}$ using only Kalman filter predictions $\mathcal{O}_{\text{pred-remain}}$. If tracks remain after this step, $\mathcal{T}_{\text{re-remain}}$, persist beyond a specific duration, $k$, without re-association, they are then discarded.

During both association phases, a cost matrix $\textbf{C}$ is formed by computing the mIOU cost between every input object detection $D_n$ and saved tracks $T^k$. Using this matrix, we solve the linear assignment problem to link each matrix row (existing tracks) with at most one unique column (object detections), adhering to the constraint
 $\min \sum_{i=1}^{\lvert \mathcal{O} \rvert} \sum_{j=1}^{\lvert \mathcal{T} \rvert} c_{ij} \cdot x_{ij} \mid x_{ij} \in \{0,1\}$, where $\lvert \mathcal{O} \rvert$ and $\lvert \mathcal{T} \rvert$ represents cardinality of $\mathcal{O}$ and $\mathcal{T}$, respectively~\cite{martello1987linear}. 

\setlength{\belowcaptionskip}{-14pt}
\begin{figure}
  \centering
  \includegraphics[width=1\linewidth]{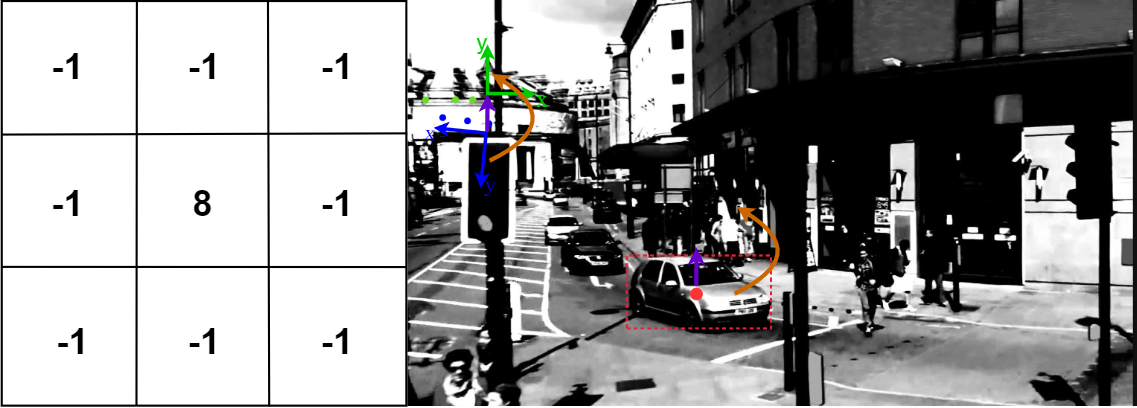}
    \vspace{-1.5em}
  \caption{\textbf{Convolution Kernel \& Transformation Visualization} On the right, blue and green points represent $\mathcal{P}_{t-1}$ and $\mathcal{P}_{t}$. Purple and orange arrows denote displacement and rotational transform. Red dashed box is the predicted object position from KF.\\}
  \label{fig：tracking}
\end{figure}
\setlength{\belowcaptionskip}{0pt}

The next step is to update current tracks $\mathcal{T}$ using associated tracks after these two associations (new locations from bounding boxes). Leftover detections from $\mathcal{O}_{\text{high}}$ trigger new tracks, with the remaining tracks removed after staying unassociated for a period. Finally, we update the previous frame keypoints $\mathcal{P}_{t-1}$ with those generated from the current frame.

For keypoints generation, we convolve the current image $I_{t}$ with a discrete approximation of a Laplacian operator, shown in Fig.\ref{fig：tracking}. The Laplacian operation, $\nabla^2 I = \frac{\partial^2 I}{\partial x^2} + \frac{\partial^2 I}{\partial y^2}$, highlights regions of rapid intensity change, such as edges and isolated points. Points with a value greater than $\theta_{th}$ remaining active, from which we sample $a$ points and save them as $\mathcal{P}_{t-1}$.

A succinct representation of C-BYTE, highlighting critical elements in \textcolor{ForestGreen}{green}, can be found in Algorithm 1.
\vspace{-0.1em}
\subsection{Trajectory Prediction}
\vspace{-0.2em}
Trajectory prediction necessitates real-time sensory data, complemented by a system skilled in identifying and tracking traffic elements. Key details such as bounding box dimensions, position, velocity, acceleration, and heading alterations are imperative for this task~\cite{leon2021review}. Central to this task is the generation of potential future scenarios $Y_\delta|\delta\in[t+1,t+m]$ informed by $n$ past coordinates $X_\eta|\eta\in[t-n,t]$. Here, each $X_\eta$ and $Y_\delta$ represents the locations of the bounding boxes (top-left and bottom-right corner) at time $\eta$ or $\delta$.

Methodologies in this domain span from Bayesian LSTMs, which utilize observation uncertainty for location predictions~\cite{bhattacharyya2018long}, to Conv1D frameworks that exploit multi-modal data for predicting pedestrian movements~\cite{yagi2018future}. Goal-driven trajectory prediction has recently gained traction, emphasizing conditional step-by-step forecasting at the cost of computational efficiency~\cite{yao2021bitrap, wang2022stepwise}.

Our design utilizes pre-trained conditional variational autoencoders to provide multi-modal information and encode past trajectories and a refined step-wise goal estimator as a decoder for future trajectories:

\setlength{\belowcaptionskip}{-14pt}
\begin{figure}
  \centering
  \includegraphics[width=1\linewidth]{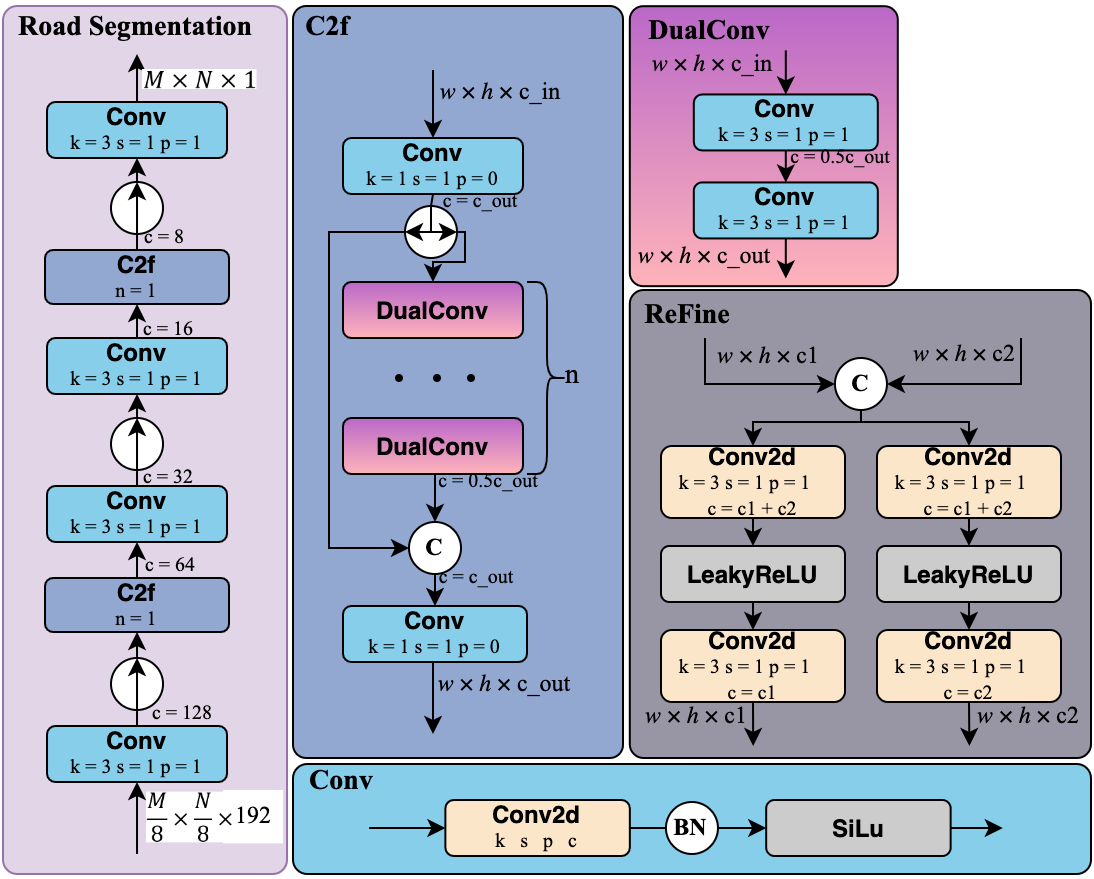}
  \caption{\textbf{Granular Model Structure.2} Design of road segmentation block, along with other components.\\}
  \label{fig：part1}
\end{figure}
\setlength{\belowcaptionskip}{0pt}

\textbf{Past Trajectory Encoder:}
When trajectories $\mathcal{T}$ generated by C-BYTE are received, they are converted to a series of past trajectories $X_\eta$ and undergo an initial encoding process via a Gated Recurrent Unit (GRU). This step refines sequential and contextual details into $h_t$ from $X_\eta$ and $h_{init}=0$. It is succeeded by a pre-trained Conditional Variational Autoencoder (CVAE), incorporated from BiTrap~\cite{yao2021bitrap}, which includes a latent prior net $P(Z|X_t)$. Based on the observed trajectories, this latent net predicts an ensemble of mean values $\mu_{Z_p}$ and covariance $\Sigma_{Z_p}$ of all conceivable future positions within the time frame $\delta\in [t+1,t+\delta]$. Then, $Z$, an array of mean and covariance samples, is drawn from $N(\mu_{Z_p},\Sigma_{Z_p})$. We then concatenate these latent variables to $h_t$, $(h_t\oplus Z)$, to inform the decoder of possible future positions in which objects can reside. Such architecture design facilitates faster extraction of salient and latent information by removing the need to rely on a succession of Recurrent Neural Networks~\cite{wang2022stepwise}. 

\textbf{Future Trajectory Decoder:}
Receiving concatenated output from the encoder, the decoder sequentially projects objects' future trajectories. Recognizing a single GRU block, generating the entire sequence of future time-step hidden states ${h_{t+1},...,h_{t+m}}$ from $h_t$, can only encapsulate data up to time $t$, we employ a sequence of discrete GRU Cells. Each individually yields $h_{t_i+1}$ from $h_{t_i}$ containing data up to time $t_i$, where $t_i\in[t,t+m-1]$. After each cell, we use a fully-connected layer to derive the final numerical location output $Y_\delta|\delta\in[t+1,t+m]$. 

During implementation, two problems arise: 1) the encoder output $(h_t \oplus Z)$, a concatenation of hidden states and latent variables, may deviate from the domain of the GRU cell's hidden state input $h$. Thus, to reduce this domain gap, we introduce another fully-connected (dense) layer to convert $(h_t \oplus Z)$ into $h_{tz}$. 2) We recognize an absence in input vector $Y_{t_i}$ for the GRU cells at future times $t_i$. And these input vectors differ from the decoded location output $Y_{t_i}$ as it must be interpretable to the GRU cell. Therefore, we append dense layers to transmute the output of the previous GRU cell, $h_{t_i-1}$, into the input vector $Y_{t_i}$.

The architecture preserves additional information, ensuring smooth data flow and enabling efficient one-directional predictions that capture comprehensive data dependencies.
\vspace{-0.2em}
\subsection{Road Segmentation}
\vspace{-0.2em}
Segmentation is a technique to divide images into regions, crucial for pinpointing and isolating objects of interest, including semantic and instance segmentation~\cite{minaee2021image}. Recent methodologies propose universal image segmentation, aiming to classify a wide range of Objects~\cite{lambert2020mseg, Yingfeng, wang2023internimage}. However, our module adopts a more focused approach, specifically designed for autonomous driving by confining the segmentation scope to drivable surfaces. The intent is to reduce the computational load necessary for high-resolution segmentation tasks.

\textbf{Segmentation block's architecture}, illustrated in Fig.\ref{fig：part1}, is minimalist but can achieve noteworthy results. Given a concentrated focus on one class, we adopt the U-Net framework, where \(O=D(E(I))\)~\cite{ronneberger2015u}. Determining our $E(I)$, we recognize that the stride-8 feature map, $\Phi_8$, from our convolutional decoder encompasses decoded data from strides 16 and 32. Decoding this feature map removes any additional processing needed for $\Phi_{16}$ and $\Phi_{32}$, enhancing processing speed. Thus, the mechanism of our block becomes $O=D(\Phi_8)$. 

\textbf{Decoder's design} aligns with a convolutional decoding blueprint, melding CBS (Conv2D-BatchNorm-SiLu) with C2f that consists of regular convolutions, DualConv, and skip connections adapted from YOLO \cite{yolov8_ultralytics} and shown in Fig.\ref{fig：part1}. Here, we replace the traditional Bottleneck block with a dual-CBS setup, further optimizing computational efficiency. By integrating these with upsampling layers, the module swiftly decodes $\Phi_8$ into a tensor $S_{h,w}$, congruent in dimensions to input $I$, representing the pixel-level road classification. This specialized approach enables our model to clearly discern drivable surfaces with faster speed than broader models. 

\subsection{Depth Estimation}

The pursuit of depth estimation from single images, crucial for robotic navigation and autonomous driving~\cite{ming2021deep}, has evolved significant, especially with the rise of deep learning. Depth learning strategies fall into three primary categorizations based on their constraints. The first strategy leverages ordinal relation constraints, employing listwise ranking mechanisms~\cite{lienen2021monocular}. The second approach emphasizes surface normal constraints, refining depth prediction borders and recognizing long-range relationships~\cite{yin2019enforcing}. Lastly, the heuristic refinement category focuses on enhancing post-prediction depth~\cite{yuan2022neural}.

Advanced methodologies often blend elements from these classifications, using variational constraints followed by refinements for precision~\cite{liu2023va}. Our module, inspired by this approach, aims to better balance computational expense with accuracy. Conceptually, depth formation mirrors single-class segmentation; the goal is to associate each pixel in an input image with a specific depth value, articulated as \(\sum_{i=1}^{M} \sum_{j=1}^{N} P(i,j,I) = D(i,j)\). Here, $M$ and $N$ represents width and height of the input image $I$. The emphasis is to devise a competent decoder, exemplified by the function $P$. In alignment with the widely adopted Encoder-Coarse-Refine methodology in monocular depth estimation~\cite{liu2023va}, we delineate the following modules for our depth estimation process:

\textbf{Coarse Depth Former:}
This module operates predominantly on condensed feature maps, establishing global depth references within images for subsequent refinements. Hence, We selected the richest layers from backbone outputs $\Phi_{16}$ and $\Phi_{32}$ for decoding purposes. To enhance processing rates, we engineered a straightforward UpDecoder, employing C2f as the primary decoding conduit for the input feature maps. An auxiliary C2f layer then formulates the preliminary depth map $D(i,j)$ for images, where $i\in [0,M/8], j\in [0,N/8]$. Such a simple configuration, therefore, offers a reduction in processing duration relative to other methods~\cite{liu2023va} while maintaining comparable efficacy.

\textbf{Refine Depth Former:}
This module's role is refining the initial depth input to produce an exact depth layer commensurate with input image $I$. To accomplish this, we once again employ U-Net, but with a tailored configuration. Instead of directly merging the upscaled depth map with the backbone feature maps $(D(i,j)\oplus\Phi_b)$ and using convolutions to form $D(i^*,j^*)$, we integrated output from a secondary flow in the refinement block. This flow has the same structure as the main flow that produces the output depth map, whose outputs are subsequently fused with the backbone feature map by the UpDecoder for further refinement. Expressed succinctly, \(D(i^*,j^*)=R(D(i,j) \oplus\Phi_{UD})\), where \(\Phi_{UD}=UD(\Phi_R \oplus \Phi_b), i^*\in[0, 2i], j^*\in[0,2j]\). Utilizing multiple such refinement layers, the final depth map $D(i_f,j_f)$ is a blend of several refined outputs $D(i^*,j^*)$, capturing details across varied scales while retaining a cohesive module structure.

\subsection{Training and Loss Function}

One challenge of training such a multi-task model is the lack of a single, comprehensive dataset that covers every module. To address this, we adopt a cross-dataset training approach. Rather than limiting our model to a singular dataset, we train individual modules on multiple datasets, each known for its strengths in specific domains. 

For instance, the Kitti dataset~\cite{geiger2013vision} specializes in monocular depth estimation and object detection. Similarly, the Cityscape dataset~\cite{cordts2016cityscapes} is used for our road segmentation module. As these modules are trained, they collaboratively refine the learnable parameters in the backbone and convolution decoder. This means that Swin transformer backbone becomes a task-agnostic module tuned on both Kitti and Cityscape datasets. On the other hand, our trajectory prediction module learns from the JAAD~\cite{kotseruba2016joint} and PIE~\cite{rasouli2019pie} datasets, which feature marked pedestrian and vehicle trajectories (Cartesian coordinates) from monocular camera videos and thus do not require the involvement of previous modules.

Our approach to integrating domain-specific losses is summarized in the equation: \(L=\lambda_{det}L_{det}+\lambda_{seg}L_{seg}+\lambda_{depth}L_{depth}+\lambda_{traj}L_{traj}\), with all parameters subject to optimization. For the weightings, we assign a value of 5 to $\lambda_{seg}$ from empirical findings, while keeping $\lambda_{det}$, $\lambda_{depth}$, and $\lambda_{traj}$ at a balanced value of 1.

\section{Experiments \& Results}
\setlength{\belowcaptionskip}{-8pt}
\begin{figure*}
  \centering
  \includegraphics[width=1\linewidth]{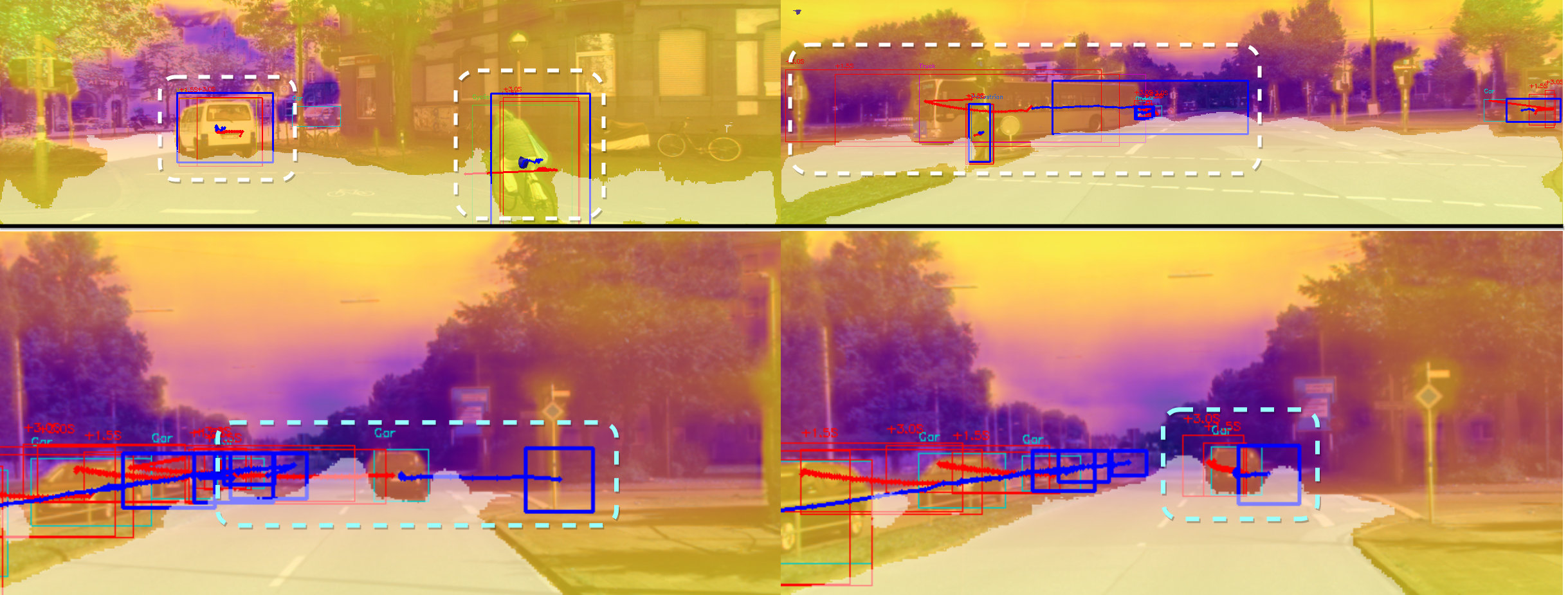}
  \vspace{-1.25em}
  \caption{\textbf{Results Visualization} The quartet of images depicts the output of LRHPerception, with past trajectories delineated in blue and future trajectory predictions in red. The \textbf{upper} pair of images exemplifies two \textbf{Success Cases}, while the \textbf{lower} duo present one \textbf{Failure Case}.}
  \label{fig：fourth}
  \vspace{-1 em}
\end{figure*}
\setlength{\belowcaptionskip}{0pt}

Our experimentation framework comprises two primary components: modular analysis to underscore the robustness of individual innovation on quantitatively testable datasets, and comprehensive assessment to exemplify our fusion techniques alongside information-rich representation of the environment. Despite parallel training, we exclusively utilize a single RTX 3090 GPU for all tests, with metrics lower the better unless stated otherwise. Bold values represent our methods, while underlined ones signify the top-performing results.

 

\setlength{\floatsep}{5pt plus 1pt minus 1pt}
\setlength{\textfloatsep}{1pt plus 1pt minus 1pt}
\setlength{\intextsep}{1pt plus 1pt minus 1pt}
\captionsetup[table]{font={small}}
{\fontsize{9pt}{10pt}\selectfont
\begin{table}[h]
\centering
\label{1}
\begin{tabular}{p{1.9cm}p{0.7cm}p{0.8cm}p{0.8cm}p{0.8cm}p{1.1cm}}
\toprule
Methods  & Detect & MOTA$\uparrow$ & IDF1$\uparrow$ & IDP$\uparrow$ & Time(ms)\\ 
\midrule


\textbf{C-BYTE} & \textbf{YoloX} & \underline{\textbf{76.9\%}} & \underline{\textbf{81.2\%}} & \textbf{85.9\%}& \textbf{31.0}\\
OC(2023)\cite{cao2023observation}         & YoloX & 74.1\% & 77.8\% & \underline{87.2}\%& 28.3\\
Byte(2022)\cite{zhang2022bytetrack}         & YoloX & 76.6\% & 79.3\% & 84.0\% & \underline{27.1}\\
BoT(2022)\cite{aharon2022bot}         &  YoloX   & 76.8\%   &  81.0\%   & 85.7\% & 48.2\\

\bottomrule
\end{tabular}
\vspace{-0.5em}
\caption{\textbf{Object Tracking.} Our model manifests a significant improvement over SOTA across all efficacy and efficiency metrics on MOT datasets, corresponding harmoniously with our structure- fundamentally predicated on achieving refined outputs for robustness. Hyperparameter $\theta_{th}$ and $a$ are empirically determined as $0.9$ and $210$.}
\vspace{-0.1 em}
\label{tab:object_tracking}
\end{table}

\begin{table}[h]
\centering
\label{2}
\begin{tabular}{p{1.9cm}p{1.7cm}p{0.6cm}p{0.7cm}p{1.8cm}}
\toprule
\multirow{-1}{*}{Methods} & \multirow{-1}{*}{MSE} & \multirow{-1}{*}{$C_{MSE}$} & \multirow{-1}{*}{$CF_{MSE}$} & \multirow{-1}{*}{FPS $\uparrow$}\\
            \textbf{Dataset: JAAD}  &      (0.5/ 1.0/ 1.5s)        &   (1.5s)      &         (1.5s)    &(8 / 12 / 24) \\
\addlinespace[-0.3em]
\midrule
\textbf{LRHP (Ours)} & \underline{\textbf{43/ 113/ 283}} & \underline{\textbf{239}} & \underline{\textbf{662}} & \textbf{\underline{111}/ 104/ 92.6}\\
SGNet (2021)          & 82/ 328/ 1049 & 995 & 4076 & 2.8/ 2.8/ 2.7\\
Bitrap (2020)          &  93/ 378/ 1206   & 1105    &  4565   &99.3/ \underline{105/ 97.1}\\
PIE\_traj (2019)       &  110/ 399/ 1280  &1183   & 4780    & -/ -/ -\\
B-LSTM (2017)       & 159/ 539/ 1535 &    1447  &   5615  & -/ -/ -\\
\addlinespace[-0.05em]
\textbf{Dataset: PIE}& & & & \\
\addlinespace[-0.3em]
\midrule
\textbf{LRHP (Ours)} & \underline{\textbf{19/ 44/ 104}} & \underline{\textbf{81}} & \underline{\textbf{233}} & \textbf{\underline{111}/ 104/ 92.6}\\
SGNet (2021)         & 34/ 133/ 442 & 413 & 1761 & 2.8/ 2.8/ 2.7\\
Bitrap (2020)          &  41/ 161/ 511   & 481    &  1949   &99.3/ \underline{105/ 97.1}\\
PIE\_traj (2019)       &  58/ 200/ 636  &596   & 2477    & -/ -/ -\\
B-LSTM (2017)       & 159/ 539/ 1535 &    1447  &   5615  & -/ -/ -\\
\bottomrule
\end{tabular}
\vspace{-0.25em}
\caption{\textbf{Trajectory Prediction.} Our model exhibits significant enhancements in all aspects of efficacy and efficiency relative to SOTA methods, particularly pronounced as prediction time expands.}
\label{tab:trajectory_prediction}
\end{table}

\begin{table}[h]
\centering
\label{3}
\begin{tabular}{p{3cm}p{1.0cm}p{0.9cm}p{1.9cm}}
\toprule
Methods  & mIOU $\uparrow$ & FPS $\uparrow$ & Remarks\\
\addlinespace[-0.3em]
\midrule
\textbf{LRHP (Ours)$^*$}  & \underline{\textbf{88.9}} & \textbf{55.0} & \multirow{-1}{*}{$^*$ speed on dual}\\
\textbf{Road Segmentation (Yolo Backbone)$^*$}  & \textbf{88.4}            & \underline{\textbf{96.4}}&  tasks of detecti-\\
InternImage (2023)\cite{wang2023internimage}          & 86.1 & 79.5 &                     on and segmen-\\               
MSeg (2020)\cite{lambert2020mseg}                  &   77.6   & 65.7  &                tation\\             
UJS-base (2021)\cite{Yingfeng}            & 80.5   &  -  &  \\
UJS-refined (2021)\cite{Yingfeng}           & 88.3  &   - & \\
\bottomrule
\end{tabular}
\caption{\textbf{Road Segmentation.} Our model exhibits an appreciable acceleration and precision enhancement when operating an additional detection task with the YOLO backbone- a testament to the robustness and effectiveness of our architectural design.}
\label{tab:road_segmentation}
\end{table}

\begin{table}[h]
\centering
\begin{tabular}{p{2.0cm}p{1cm}p{0.8cm}p{0.8cm}p{0.8cm}p{0.7cm}}
\toprule
Methods  & Backbone & RMS & $\delta_1$ $\uparrow$  &  $\delta_2$ $\uparrow $ & FPS $\uparrow$\\
\addlinespace[-0.3em]
\midrule
\textbf{LRHP (Ours)} & \textbf{Swin-m} & \textbf{0.229} & \textbf{0.966} &\textbf{0.996} & \underline{\textbf{42.0}}\\
\textbf{Depth Estimate} & \textbf{Swin-L} & \textbf{0.216} & \textbf{0.975}& \underline{\textbf{0.997}} & \textbf{13.3} \\
VA-Depth (2023)         & Swin-L & \underline{0.209} & \underline{0.977} & \underline{0.997} &6.2\\
AdaBins (2021)      &  EffNet\&Vit   & 0.236  &  0.964   & 0.995 & 1.7\\
BTS (2019)         &  DenseNet   & 0.280   &  0.955   &0.993   & 20.1\\
ASTrans (2021)         &  Vit-B   & 0.269   &  0.963   &0.995 & -\\
DORN (2018)         &  ResNet   & 0.273   &  0.932   &0.984    & -\\
\bottomrule
\end{tabular}
\vspace{-0.3em}
\caption{\textbf{Depth Estimation.} Our model showcases a significant acceleration in processing speed relative to SOTA, all the while maintaining comparable accuracy levels, thus underscoring the remarkable efficiency of our design.}
\label{tab:depth_estimation}
\end{table}}

\subsection{Modular Results}
For modular analysis, a complete cross-dataset trained package is tested on all tasks except object tracking. Comparison models are singularly trained for their specific tasks. As our C-BYTE for object tracking lacks learnable parameters, we employ ByteTrack's framework as the benchmarking standard.
\vspace{-1em}
\subsubsection{\textbf{Object Tracking}}
\vspace{-0.3em}

We evaluate C-BYTE within ByteTrack's framework to use the same backbone of YoloX-x, following the ``private detection" protocol across MOT17 (multiple-object-tracking) datasets~\cite{milan2016mot16}. We employ the conventional metrics of MOTA (Multiple Object Tracking Accuracy), IDF (ID F1 Score), and IDP (ID Precision), to assess various facets of tracking: MOTA is based on FP (false positive), FN (false negative) and IDs, while IDF and IDP evaluate association performance by penalizing inaccurate tracks.

The resultant data in Table.\ref{tab:object_tracking} exhibits a noticeable advancement over Byte and other SOTA methods, substantiating that camera motion correction refined the Kalman Filter's predictions by removing the nonlinear disturbance for which linear models of KF cannot account. With a negligible delay of less than 4 milliseconds compared to Byte, our method demonstrates superior tracking results across metrics. The sturdiness of this module is further validated in the visualizations of joint tests, where we observe minimal to nonexistent instances of failure attributable to the trajectory former. This empirical evidence corroborates the efficacy and reliability of our model.
\vspace{-0.4em}

\subsubsection{\textbf{Trajectory Prediction}}
\vspace{-0.2em}

Our prediction former is assessed on JAAD and PIE datasets~\cite{kotseruba2016joint,rasouli2019pie}, which feature ego-centric videos annotated at 30Hz. Following established benchmarks~\cite{rasouli2019pie}, we utilize a 15-frame observational duration and a 45-frame prediction horizon for the evaluation, where the ground-truth observation is given. We additionally introduce a pragmatic speed assessment wherein we aggregate tracks in batches of 8, 12, and 24, indicative of the object count within a single image, and ascertain how many batches can be processed per second.

The outcomes in Table.\ref{tab:trajectory_prediction} evince a distinct augmentation in both speed and accuracy across both datasets, a discrepancy that broadens as the prediction timeline extends into the future. In quantitative terms, our model presents an impressive \textbf{40-fold} increase in processing speed compared to the alternate highest-accuracy method\cite{wang2022stepwise} and a \textbf{4-fold} boost in accuracy over the quickest model\cite{yao2021bitrap}. Such robust performance attests to the efficacy of an encoder capturing explicit and latent dependency and a decoder featuring swift unidirectional prediction.
\vspace{-0.4em}

\subsubsection{\textbf{Road Segmentation}}
\vspace{-0.2em}

Our segmentation former is evaluated on Cityscape dataset~\cite{cordts2016cityscapes}. The dataset encompasses pixel segmentation across all classes present, with our module specifically targeting drivable surfaces. Tests are conducted on the validation set with models presenting mIOU (mean intersection over union) across the entire set. During speed evaluation, we execute both detection and segmentation as our module is integrated within a convolution decoder for object detection. We also implement a standalone segmentation module using the YOLO backbone to illustrate the simplicity of the segmentation module itself.

Remarkably, our module surpasses the performance of prevailing universal modules in Table.\ref{tab:road_segmentation}, underscoring the potency of simplicity in certain domains. Although our final choice of the Swin backbone, geared towards combined modules, increases segmentation accuracy but reduces processing speed, this module still excels in executing its designated task in subsequent joint tests, contributing to the overall goal of real-time monocular perception.
\vspace{-0.4em}

\subsubsection{\textbf{Depth Estimation}}
\vspace{-0.2em}

Our depth estimator is assessed on the KITTI dataset~\cite{geiger2013vision} adhering to the splits delineated in ~\cite{liu2023va}, where depth maps are annotated with a range of 0 to 80 meters. A selection of quintessential metrics, including RMS (root mean square error) for overall error estimation and $\delta_1$ and $\delta_2$ for precision within specific tolerances, are employed. To underscore the efficacy of our module design, we also implement a stand-alone depth estimator with the Swin-L backbone, neutralizing the computational variances on backbones to evaluate our depth estimation decoder design.

The yielded results in Table.\ref{tab:depth_estimation} underscore a noteworthy acceleration in frames processed per second, whilst maintaining a high degree of accuracy. Concretely, our design manifests a \textbf{577\%} uplift in processing speed over the best-alternative\cite{liu2023va}. Should the same backbone be used, our decoder design alone offers \textbf{115\%} improvement in speed with comparable accuracy. These values validate the design of a simplified coarse-refine layout using modified C2f layers. The success of this module affords our suite the ability to produce accurate vital depth information without sacrificing real-time capabilities.
\vspace{-0.4em}
\subsubsection{\textbf{Ablation Study}}
\vspace{-0.2em}

We studied the choice of hyperparameter $n$ in our introduction of the C2f module into road segmentation and depth estimation, presented in Table.\ref{tab:ablation_study}. With this study, we incorporated two DualConv modules in the C2F to balance the efficacy and computational need.

{\fontsize{9pt}{10pt}\selectfont
\begin{table}[h]
\centering
\begin{tabular}{p{1.4cm}p{1.5cm}p{1.7cm}p{2cm}}
\toprule
Number of DualConv  & Road-Seg (mIOU)$\uparrow$ & Depth Estimation (RMS) & Time Difference ($T_n \!-\!T_{n=1}$)(ms) \\
\addlinespace[-0.3em]
\midrule
\addlinespace[+0.1em]
$n=1$ & \underline{88.5} & 0.223 & \underline{0}\\
$n=3$ & \underline{88.5} & \underline{0.219} & 0.268\\
\addlinespace[-0.2em]
\midrule
\addlinespace[0.2em]
\textbf{$n=2$} & \textbf{88.9} & \textbf{0.220} & \textbf{0.049}\\
\addlinespace[-0.3em]
\bottomrule
\end{tabular}
\vspace{-0.3em}
\caption{\textbf{Ablation study.} We studied the effect of different numbers of Dual Convolutions in the C2f module on depth estimation and road segmentation.}
\label{tab:ablation_study}
\vspace{-0.3 em}
\end{table}}

{\fontsize{9pt}{10pt}\selectfont
\subsection{Joint Results}
\begin{table}[h]
\centering
\begin{tabular}{p{3cm}p{3cm}p{1cm}}
\toprule
Methods  & Category & FPS $\uparrow$\\
\addlinespace[-0.3em]
\midrule
\textbf{LRHP(Ours)} & \underline{\textbf{Monocular}} & \underline{\textbf{28.8} }\\
LRHP in series        & Monocular   & 16.3\\
SOTA in series        &Monocular   & 1.8\\
Uni-AD (2023)\cite{hu2023planning}         &  Multi-Cam Map  & 2.1\\
BEVerse-Tiny (2022)\cite{zhang2022beverse}         &  Multi-Cam Map   & 4.4\\
DETR3D (2021)\cite{wang2022detr3d}          & Multi-Cam Map &2.0   \\
\bottomrule
\end{tabular}
\vspace{-0.1em}
\caption{\textbf{Entire Model.} Our model witnesses an improvement of more than an order of magnitude over existing local mapping methods, a testament to the exceptional efficiency of our model. Note that Uni-AD's planning module was removed for a fair comparison.}
\label{tab:entire_model}
\vspace{-0.0 em}
\end{table}}

We carry out comprehensive assessments on KITTI Dataset ~\cite{geiger2013vision}, featuring videos captured at a rate of 10 frames per second. This selection of the dataset serves to highlight the practical efficacy of our model under real-world scenarios, bolstering its applicability and value for future research.

\textbf{Quantitative.} In our empirical comparison, we juxtapose the computational demands of our method against multi-camera map techniques and monocular methods in Table.\ref{tab:entire_model}. The latter are constructed from the current SOTA solutions in each domain of tracking, trajectory prediction, road segmentation, and depth estimation. Remarkably, LRHPerception facilitates a real-time processing rate of \textbf{29} FPS, constituting a substantial \textbf{555\%} acceleration over the fastest mapping technique. Upon scrutinizing the contributions to this efficiency, our module enhancements account for an 806\% acceleration relative to sequentially-connected SOTA methods, with our integration technique further doubling the speed-up to \textbf{1500\%}.

\textbf{Qualitative.} Given that our perception package represents an unparalleled fusion of functionalities, we resort to visualization for qualitative assessment. Fig.3 encapsulates successful instances, embodied in white boxes, including a right-turning van and a forward-moving bicycle along a road and a left-turning bus and a stationary pedestrian in an intersection. Recognizing the pedagogical value of shortcomings, we also feature failure cases. One typical scenario illustrates a right-turning car mispredicted to continue leftward, with the correct forward trajectory identified half a second later. Furthermore, the segmentation module overlooks a potential route to the right of the intersection. These areas of discrepancy delineate the LRHPerception for future enhancements.

\section{Conclusion}

In this work, we unveil LRHPerception, a monocular perception package that achieves a balanced information richness and computational load. It efficiently and seamlessly blends road identification, object surveillance, trajectory prediction, and distance approximation, aligning ego-planner with human perception, achieving real-time functionality while offering human-understandable interpretation to the surrounding environment. Serving as a humble groundwork for exploration, LRHPerception constitutes an efficient toolkit, laying work for future ingenuity in the safe comprehensible autonomous driving domain.

\vspace{-1em}

\bibliographystyle{IEEEtran}
\bibliography{ref}
\end{document}